\documentclass{article}
\pdfoutput=1
\usepackage{spconf,amsmath,graphicx}
\usepackage{enumitem}
\usepackage{graphicx}
\usepackage{subcaption}
\usepackage{lpic}
\usepackage{microtype}
\usepackage{cite}
\usepackage{booktabs}

\title{RNN TRANSDUCER MODELS FOR SPOKEN LANGUAGE UNDERSTANDING}
\name{\begin{tabular}{c} Samuel Thomas, Hong-Kwang J. Kuo, George Saon, Zolt{\'a}n T\"{u}ske, \\Brian Kingsbury, Gakuto Kurata, Zvi Kons, Ron Hoory\end{tabular}}
\address{IBM Research AI}

\begin{document}
\ninept
\sloppy
\maketitle
\begin{abstract}
We present a comprehensive study on building and adapting RNN transducer (RNN-T) models for spoken language understanding (SLU). These end-to-end (E2E) models are constructed in three practical settings: a case where verbatim transcripts are available, a constrained case where the only available annotations are SLU labels and their values, and a more restrictive case where transcripts are available but not corresponding audio.  We show how RNN-T SLU models can be developed starting from pre-trained automatic speech recognition (ASR) systems, followed by an SLU adaptation step. In settings where real audio data is not available, artificially synthesized speech is used to successfully adapt various SLU models. When evaluated on two SLU data sets, the ATIS corpus and a customer call center data set, the proposed models closely track the performance of other E2E models and achieve state-of-the-art results. 
\end{abstract}
\section{Introduction}
\label{sec:intro}

Recently there has been a significant effort to build end-to-end (E2E) models for spoken language understanding~\cite{serdyuk2018towards,Haghani2018,qian2017exploring,chen2018spoken,ghannay2018end,lugosch2019speech,caubriere2019curriculum,huang2020leveraging,lugosch2020using,price2020improved,radfar2020end,tian2020improving,jia2020large,kuo2020end,palogiannidi2020end}. 
Instead of using an ASR system in tandem with a text-based natural language understanding system~\cite{Goel2005,yaman2008integrative,Haghani2018}, these systems directly process speech to produce spoken language understanding (SLU) entity or intent label targets.
Many of these end-to-end SLU systems are  trained on speech utterances with verbatim transcripts annotated additionally with SLU labels. In  a \textbf{[FULL]} setting for example, speech data is available with transcripts annotated with various SLU labels ((2) and (6) below): entities such as \mbox{\scriptsize B-toloc.city\_name} and an intent label like \mbox{\scriptsize INT-FLIGHT}. Entity labels paired with corresponding values represent important keywords conveying specific information; intent labels map entire utterances to class labels indicating overall sentence meaning. 

In practical SLU settings, these annotations may be limited in multiple ways, as illustrated below. We consider an  \textbf{[AUDIO]} setting, where audio recordings are available, but the annotations are just SLU entity label/value pairs and intents ((3)--(5)). In example (3), the entities are available, but without verbatim transcripts. Similar to (3), in (4) the SLU annotations are available, but the label/value pairs are sorted alphabetically by label.  This setting corresponds to the semantic frame or bag of entities concept where the order of entities does not affect the meaning. In a more limited setting like (5), neither transcripts nor entities are available; instead, only a single intent label is provided for each utterance. Another practical setting is \textbf{[TEXT]}, where transcripts with SLU annotations are available as in (2) and (6), but the corresponding human speech recordings are not, due to privacy restrictions or bootstrapping from text chat data.

\begin{description}[leftmargin=*]
    \item[{\bf (1) Transcript:}] {\it i want a flight to dallas from reno that makes a stop in las vegas}
    \item[{\bf (2) Transcript + Entity labels:}] {\it i want a flight to} DALLAS {\scriptsize \mbox{B-}toloc.city\_name} {\it from} RENO \mbox{\scriptsize B-fromloc.city\_name} {\it that makes a stop in} LAS \mbox{\scriptsize B-stoploc.city\_name} VEGAS \mbox{\scriptsize I-stoploc.city\_name}
    
    \item[{\bf (3) Entities in spoken order:}] DALLAS \mbox{\scriptsize B-toloc.city\_name} RENO \mbox{\scriptsize B-fromloc.city\_name} LAS \mbox{\scriptsize B-stoploc.city\_name} VEGAS \mbox{\scriptsize I-stoploc.city\_name}
    
    \item[{\bf (4) Entities in alphabetic order:}] RENO \mbox{\scriptsize B-fromloc.city\_name} LAS \mbox{\scriptsize B-stoploc.city\_name} VEGAS \mbox{\scriptsize I-stoploc.city\_name} DALLAS \mbox{\scriptsize B-toloc.city\_name}
    
    \item[{\bf (5) Intent label only:}] \mbox{\scriptsize INT-FLIGHT}
    
    \item[{\bf (6) Transcript + Intent label:}] {\it i want a flight to dallas from reno that makes a stop in las vegas} \mbox{\scriptsize INT-FLIGHT}
\end{description}

In contrast to other E2E based SLU systems in the literature, in this paper we investigate various methods to train a different class of E2E models in the various settings described above: RNN-T based SLU systems. RNN-T models typically consist of three different sub-networks: a transcription network, a prediction network, and a joint network~\cite{graves2012sequence}. The transcription network produces acoustic embeddings, while the prediction network resembles a language model in that it is conditioned on previous non-blank symbols produced by the model. The joint network combines the two embedding outputs to produce a posterior distribution over the output symbols. This architecture elegantly replaces a conventional ASR system composed of separate acoustic model, language model, pronunciation lexicon, and decoder components, using a single end-to-end trained, streamable, all-neural model that has been widely adopted for speech recognition~\cite{he2019streaming,rao2017exploring,li2019improving,shafey2019joint,ghodsi2020rnn}. 
Given their popularity, and the fact that RNN-T models can naturally handle more abstract output symbols such as ones marking speaker turns~\cite{shafey2019joint}, in this paper we explore the extension of these models to SLU tasks, building on recent advances with RNN-T models for speech recognition as described in \cite{george2021rnn}.

\section{Design considerations for SLU models}
\label{sec:adaptation}

Since the amount of annotated SLU training data is usually limited (typically only a few tens of hours of data), end-to-end models for SLU tasks are often bootstrapped from pre-trained models developed for ASR. When an ASR model is used to bootstrap an SLU system, a key design question is how accurate the pre-trained ASR model should be.
If an SLU model is bootstrapped from an ASR model trained on just a few tens of hours, how much impact on performance will that have, compared to bootstrapping from a more accurate ASR model trained on more data? 
Since verbatim transcripts are available in the \textbf{[FULL]} setting, it is possible to adapt the pre-trained ASR model prior to using it to initialize the model for SLU training. Is such an explicit ASR based adaptation step necessary or does the model implicitly adapt to linguistic content while being adapted as an SLU model? Understanding the impact of ASR adaptation is also relevant for the \textbf{[AUDIO]} setting because full transcripts are not available and hence an explicit ASR adaptation step is not possible. In such cases, can a useful RNN-T based SLU model be constructed with only partial transcripts or even without any transcripts? 

While it might seem that the \textbf{[TEXT]} case, which assumes no audio data, is extreme, it is in fact a very common SLU setting.  It is becoming increasingly common that speech is considered personally identifiable information which must be protected; thus, in some cases only transcripts of what was said will be retained instead of the audio recording.   In such scenarios, is it possible to use synthetic speech data created using a text-to-speech system to serve as a surrogate for the actual audio to effectively train an RNN-T based model? Compared to graphemes or phonemes that form the ASR output symbol set, SLU entity and intent labels used in all these settings are symbols that do not map directly to the underlying acoustic signal. When an ASR pre-trained model is used, although the model is not explicitly trained to predict non-acoustic symbols, it learns to produce these symbols in the SLU adaptation step. Given that pre-training is an important step, should the ASR pre-training step also include non-acoustic symbols? 
We construct various RNN-T based models and run experiments to address these and other questions.

\section{Building RNN-T based SLU models}

To build various kinds of SLU models we elaborate on four key components of the adaptation pipeline. These include:
\begin{enumerate}[wide, labelwidth=!, labelindent=0pt]
\item \textbf{Pre-trained ASR models} As described earlier, a key design consideration when using a pre-trained ASR model is how well trained and accurate the pre-trained ASR model is. To understand this better we train ASR models on varying amounts of task independent data. The models trained on only few tens of hours of data will have much lower ASR performance than models trained on hundreds of hours of data, but the implications for SLU performance are less clear.
\item \textbf{ASR and SLU adaptation} Starting from the pre-trained ASR models, we then investigate the impact of ASR adaptation using domain-specific speech data prior to SLU transfer learning. This kind of adaptation is possible in the \textbf{[FULL]} use case where verbatim transcripts are available. After ASR adaptation of the model, SLU labels are introduced into the training pipeline along with the full transcripts. This step is a form of curriculum learning~\cite{bengio2009curriculum,caubriere2019curriculum} that gradually modifies an off-the-shelf ASR model into a  domain-specific SLU model. What is novel in this step is that the model is now trained to output non-acoustic entity and intent tokens in addition to the usual graphemic or phonetic output tokens for ASR.  In the pre-training step, the targets are graphemic/phonetic tokens only, whereas for the SLU adaptation, the targets also include semantic labels.  Starting with an ASR model, the new SLU labels are integrated by modifying the joint network and the embedding layer of the prediction network to include additional output nodes. The new network parameters are randomly initialized, while the remaining parts are initialized from the pre-trained network.
In contrast, in the \textbf{[AUDIO]} and \textbf{[TEXT]} cases, a prior ASR adaptation step cannot be performed because full transcripts or audio are not available. The pre-trained ASR model in these cases is directly adapted into an SLU model. 
\item \textbf{SLU adaptation with synthesized speech} As described earlier, the RNN-T model has three sub-networks. We hypothesize that during the SLU adaptation process, the transcription network gets adapted the most to accommodate changes in the acoustic space from new speakers and domain mismatches. On the other hand, the addition of new non-acoustic SLU output symbols leads to the greatest changes in the prediction and joint networks. To include new additional target nodes for SLU, the joint network is modified as described earlier: new network parameters are randomly initialized while the remaining parts of the network are copied from a pre-trained network. In the \textbf{[TEXT]} case, given that only transcripts are available, we adapt the RNN-T network with synthesized speech generated using a text-to-speech system. In this adaptation process, instead of adapting the entire RNN-T model, it is possible to modify just the prediction and joint networks while keeping the transcription network fixed. This protects the transcription network from over-tuning to a synthetic voice while still learning from the SLU text data.
\item \textbf{Pre-training with pseudo-SLU tags} Although ASR pre-training suitably initializes the SLU model to process text that is related to the underlying acoustics, it is not obvious if this pre-training helps the network recognize SLU entity and intent labels.  We hypothesize that a suitable pre-training for this can be done by training ASR models not only on lexical symbols but also on part-of-speech (POS) tags which can be assigned to each word in the transcript. Similar to the SLU labels, these POS tags are also non-acoustic symbols and hence including them in pre-training could be useful to help the model learn to predict such symbols.  
\end{enumerate}

\section{Evaluating RNN-T based SLU models}
\label{sec:experiments}

\subsection{SLU Data and Evaluation Metric}

Using the training methods outlined above, we develop various RNN-T based end-to-end SLU systems on two SLU data sets.

\begin{enumerate}[wide, labelwidth=!, labelindent=0pt]
\item \textbf{ATIS} In our first set of experiments we use the ATIS \cite{hemphill1990atis} training and test sets: 4976 training utterances from Class A (context independent) training data in the ATIS-2 and ATIS-3 corpora and 893 test utterances from the ATIS-3 Nov93 and Dec94 data sets.  The 4976 training utterances comprise $\sim$9.64 hours of audio from 355 speakers. The 893 test utterances comprise $\sim$1.43 hours of audio from 55 speakers. The data was originally collected at 16~kHz, but we downsampled to 8~kHz to better match telephony use cases where we could use off-the-shelf ASR models trained on conversational telephone speech. To better train the proposed E2E models, additional copies of the corpus are created using speed/tempo perturbation~\cite{Ko15}. The training corpus after data augmentation is $\sim$140 hours of data \cite{kuo2020end}.
The ATIS task includes both entity (slot filling) and intent recognition.

\item \textbf{Call Center (CC)} The second data set we use for our experiments are based on an internal data collection consisting of call center recordings of open-ended first utterances by customers describing the reasons for their calls~\cite{Goel2005}.  The 8kHz telephony speech data was manually transcribed and labeled with correct intents.  The corpus contains real, spontaneous utterances from customers, not crowd-sourced scripted or role-played data, and it includes a wide variety of ways that customers naturally described their intents. The training data consists of 19.5 hours of speech that was first divided into a training set of 17.5 hours and a held-out set of 2 hours.  A separate data set containing 5592 sentences (5h, 40K words) was used as the final test set~\cite{huang2020leveraging}.  This task contains only intent labels and does not have any labeled semantic entities.
\end{enumerate}

We measure slot filling performance on the ATIS data set with the F1 score as in our previous work \cite{kuo2020end}. When using speech input instead of text, word errors can arise. The F1 score requires that both the slot label and value must be correct.  The scoring ignores the order of entities, and is therefore suitable for the ``bag-of-entities'' case as well. On both the ATIS and CC data sets we also measure intent recognition accuracy. 
Our intent recognition experiments are performed in \textbf{[AUDIO]} setting, where we train on only the intent label without any word transcript (5); intent accuracy is similar when trained on transcripts+intent labels (6).

\subsection{ASR pre-training}

As described earlier, the RNN-T models we develop for SLU are first pre-trained on task independent ASR data.  To train these ASR models we use three different amounts of data: 64 hours, 160 hours, and 300 hours of data from the Switchboard (SWB) corpus. RNN-T models are built on these data sets using the steps described in \cite{george2021rnn}. The training process starts by training CTC acoustic models that are used to initialize the  transcription network of the RNN-T models. We use the recipe steps presented in \cite{audhkhasi2019forget,kurata2019guiding} to construct the three different CTC based models. All the pre-trained RNN-T models we use for our experiments have a transcription network which contains 6 bidirectional LSTM layers with 640 cells per layer per direction. The prediction network is a single unidirectional LSTM layer with only 768 cells. The joint network projects and combines the 1280-dimensional stacked encoder vectors from the last layer and the 768-dimensional prediction net embedding to 256 dimensions. After the application of hyperbolic tangent, the output is projected to 46 logits, corresponding to 45 characters plus BLANK, followed by a softmax layer. The models were trained in PyTorch on V100 GPUs for 20 epochs using SGD variants with two optimizers, two learning rate schedules and two batch sizes.  More details of these design choices can be found in \cite{george2021rnn}. As described earlier, during SLU adaption, new network parameters are randomly initialized while the remaining parts of the network are copied from the pre-trained  network.  For the ATIS entity/intent task we add 151 extra output nodes to the pre-trained network as entity/intent targets. For the Call Center intent task, 29 additional nodes are used.

\begin{table}[htpb]
    \caption{ASR WER performance before and after SLU adaptation. \textbf{[P]} denotes experiments focused on pre-training.}
    \vspace{-2mm}
	\centering
	\begin{tabular}{@{}ccc@{}} \toprule
	\multicolumn{1}{c}{} & \multicolumn{1}{c}{\bf PT. Data (Hrs.)} & {\bf ATIS (WER\%)} \\ \midrule
	\textbf{[1P]} &  0   &  14.8 \\
	\textbf{[2P]} &  64  & 38.3 $\rightarrow$ 2.2 \\
	\textbf{[3P]} &  160 & 18.6 $\rightarrow$ 1.8 \\
	\textbf{[4P]} &  300 & 13.1 $\rightarrow$ 1.6 \\
	\bottomrule
	\end{tabular}
	\label{tab:asr_atis}
\end{table}

\subsection{SLU models in the \textbf{[FULL]} setting}
\label{subsec:slufull}

In the \textbf{[FULL]} setting, in addition to audio data and SLU labels, verbatim transcripts are also available to train SLU systems. We use the entity recognition task on the ATIS corpus as an example of this setting. SLU models are trained, starting from the pre-trained ASR models, on ATIS data (see example 2, Section~\ref{sec:intro}). As described earlier, one of the key design considerations in this setting is to understand how accurate the pre-trained ASR models need to be. Experiments \textbf{[1P]}-\textbf{[4P]} in Table \ref{tab:asr_atis} show the WER performance of the various pre-trained models we employ. 
Error rates were measured after filtering out entity/intent labels.
The performance of pre-trained ASR models on the ATIS test set is correlated with the amount of training data used in pre-training, ranging from 38.3\% to 13.1\%.  SLU adaptation significantly improves the WER, e.g. from 38.3\% to 2.2\% for the 64 hour pre-trained model, and brings all the models to a similar operating range (2.2\% to 1.6\%).  Note that these experiments did not involve a separate ASR adaptation step.  Thus the SLU training, given full transcripts and semantic annotations, is also able to implicitly adapt to the word content, resulting in significant WER improvement.

We further compare the performance of these models on the ATIS entity recognition task in Table \ref{tab:slu_atis_cc}. From experiments \textbf{[5P]}-\textbf{[8P]}, it is clear that while pre-training is necessary, the initial performance of the pre-trained model is not critical. All three pre-trained models perform about 12\% (e.g. 92.1\%-79.7\%) better than a model without any pre-training in terms of F1 score for entity recognition. Given the similarity in model performance, we use the 160 hour pre-trained model for all subsequent experiments.

Since full transcripts are available in this setting, we also investigate the effect of an intermediate ASR adaptation step prior to SLU adaptation. After ASR pre-training (ASR PT), the RNN-T model is adapted with just available word transcripts (ASR Adapt). This is followed by transfer learning of SLU training where the model is adapted on the same audio data with words and semantic labels. In this step, new output nodes to represent entity labels are added. Experiment \textbf{[3F]} of Table \ref{tab:entity_atis} shows just a slight improvement with this additional step, compared with experiment \textbf{[2F]} where the model is directly adapted to the SLU task. We hypothesize that when a model is adapted to an SLU task with full transcripts, it also implicitly adapts as an ASR system. In a second set of experiments, we check if pre-training the RNN-T on not only task independent ASR data but also task independent non-acoustic labels like POS tags is beneficial. We train a separate RNN-T model on 160 hours of SWB data along with machine generated POS tags generated using the SIRE toolkit~\cite{florian2004statistical}. This model is then adapted to the SLU task in experiment \textbf{[4F]}. We do not observe any gains on entity recognition with this additional step. Overall, the RNN-T models perform on par with previous attention based models developed in \cite{kuo2020end}.

\begin{table}[htpb]
    \caption{SLU performance with various pre-trained models.}
    \vspace{-2mm}
	\centering
	\begin{tabular}{@{}ccccc@{}} \toprule
	\multicolumn{1}{c}{} & \multicolumn{1}{c}{\bf PT.} & \multicolumn{2}{c}{\bf ATIS} &  \multicolumn{1}{c}{\bf CC} \\ 
	\multicolumn{1}{c}{} & \multicolumn{1}{c}{\bf Data (Hrs.)} & \multicolumn{1}{c}{\bf Ent. (F1)} & \multicolumn{1}{c}{\bf Int. (Acc\%)} &  \multicolumn{1}{c}{\bf Int. (Acc\%)} \\ 
	\midrule
	\textbf{[5P]} &  0   &  79.7 & 83.5 & 65.8 \\
	\textbf{[6P]} &  64  & 92.1 & 95.4 &  86.9  \\
	\textbf{[7P]} &  160 & 93.2 & 94.7 &  87.4 \\
	\textbf{[8P]} &  300 & 93.2 & 94.9 &  87.4 \\
	\bottomrule
	\end{tabular}
	\label{tab:slu_atis_cc}
\end{table}

\subsection{SLU models in the \textbf{[AUDIO]} setting}

In this setting, the speech utterances for SLU training are annotated only with entity label/value pairs and/or intent labels, and the entity label/value pairs may not be in spoken order. In some cases, only a single intent label that represents the \textit{meaning} of the entire utterance is available (see examples (3)-(5), Section~\ref{sec:intro}). 

First, like the experiments in the \textbf{[FULL]} setting in Section~\ref{subsec:slufull}, we conduct experiments to  measure the sensitivity of the SLU model in terms of intent classification to the initial ASR performance of the pre-trained models used to bootstrap the systems.
Regarding the intent classification experiments \textbf{[5P]}-\textbf{[8P]} in Table~\ref{tab:slu_atis_cc} on ATIS, our observations are consistent with the entity recognition results:  pre-training is important, but the initial ASR performance of the pre-trained model is not critical. All three pre-trained models perform up to 12\% (e.g.~95.4\%-83.5\%) better than a model without pre-training in terms of intent recognition accuracy.
We observe similar results on the CC data set as well. In this case we observe about 20\% (87.4\%-65.8\%) improvement in intent accuracy when a pre-trained model is used over training an SLU model from scratch.

\begin{table}[t]
    \caption{SLU entity recognition performance on ATIS in various settings denoted by tags: \textbf{[F]} denotes \textbf{[FULL]}, \textbf{[A]} denotes \textbf{[AUDIO]}, and \textbf{[T]} denotes \textbf{[TEXT]}.} 
    \vspace{-2mm}
	\centering
	\begin{tabular}{@{}ccc@{}} \toprule
	\multicolumn{1}{c}{} &  \multicolumn{1}{c}{\bf Model Building} & \multicolumn{1}{c}{\bf Ent. (F1)} \\ 
	\midrule
	\textbf{[1F]} & Attention-based model\cite{kuo2020end} & 93.0 \\
	\textbf{[2F]} & ASR PT $\rightarrow$ SLU Adapt & 93.2 \\
	\textbf{[3F]} & ASR PT $\rightarrow$ ASR Adapt $\rightarrow$ SLU Adapt & 93.7 \\
	\textbf{[4F]} & ASR+POS PT $\rightarrow$ SLU Adapt & 93.2 \\
	\midrule
	\textbf{[1A]} & ASR PT $\rightarrow$ SLU Adapt (s-order) & 93.9 \\
	\textbf{[2A]} & ASR PT  $\rightarrow$ SLU Adapt (a-order) & 79.4 \\
	\midrule
	\textbf{[1T]} & ASR PT $\rightarrow$ SLU Adapt-Real (PR+Joint)$^*$ & 90.8 \\
	\textbf{[2T]} & ASR PT $\rightarrow$ SLU Adapt-SS (PR+Joint) & 90.7 \\
	\textbf{[3T]} & ASR PT  $\rightarrow$ SLU Adapt-SS (ALL) & 91.6 \\
	\textbf{[4T]} & ASR PT  $\rightarrow$ SLU Adapt-MS (ALL) & 93.2 \\
	\bottomrule
	\end{tabular}
	\label{tab:entity_atis}
\end{table}

We next conduct an oracle experiment (marked with *) (\textbf{[5A]}) in which we perform ASR adaptation before SLU training on only intent labels: this is an oracle experiment because we assume transcripts are unavailable for SLU training. A comparison of experiments \textbf{[4A]} and \textbf{[5A]} in Table~\ref{tab:intent_atis} shows that the model performs reasonably well without any transcripts in training to recognize intent labels, but would benefit if transcripts were available.
We observe similar results with intent recognition on the Call Center data set (compare \textbf{[8A]} to \textbf{[9A]} in Table \ref{tab:intent_cc}; \textbf{[9A]} is an oracle experiment assuming transcripts were available). Using POS markers as pseudo-SLU targets has mixed results: while we observe gains on the ATIS task (compare \textbf{[4A]} to \textbf{[6A]}; Table \ref{tab:intent_atis}), there are none on the Call Center task (compare \textbf{[8A]} to \textbf{[10A]} in Table \ref{tab:intent_cc}).

\begin{table}[htpb]
    \caption{SLU intent recognition performance on ATIS in various settings}
    \vspace{-2mm}
	\centering
	\begin{tabular}{@{}ccc@{}} \toprule
	\multicolumn{1}{c}{} &  \multicolumn{1}{c}{\bf Model Building} & \multicolumn{1}{c}{\bf Int. (Acc\%)} \\ 
	\midrule
	\textbf{[3A]} & Attention-based model\cite{kuo2020end} & 96.9 \\
	\textbf{[4A]} & ASR PT $\rightarrow$ SLU Adapt & 94.7 \\
	\textbf{[5A]} & ASR PT $\rightarrow$ \textit{ASR Adapt} $\rightarrow$ SLU Adapt$^*$  & 96.3 \\
	\textbf{[6A]} & ASR+POS PT $\rightarrow$ SLU Adapt & 95.9 \\
	\midrule
	\textbf{[5T]} & ASR PT $\rightarrow$ SLU Adapt-Real (PR+Joint)$^*$ & 93.7 \\
	\textbf{[6T]} & ASR PT $\rightarrow$ SLU Adapt-SS (PR+Joint) & 93.1 \\
	\textbf{[7T]} & ASR PT  $\rightarrow$ SLU Adapt-SS (ALL) & 93.8 \\
	\textbf{[8T]} & ASR PT  $\rightarrow$ SLU Adapt-MS (ALL) & 95.4 \\
	\bottomrule
	\end{tabular}
	\label{tab:intent_atis}
\end{table}

Similar to the intent experiments, we conduct entity recognition experiments on the ATIS data set without verbatim transcripts. The models in this case are trained with just the entity labels and their corresponding values. We experiment with two ways of ordering the entities: in spoken order and also alphabetical order (see examples (3)-(4), Section~\ref{sec:intro}). Experiments \textbf{[1A]} and \textbf{[2A]} of Table \ref{tab:entity_atis} show the results in these settings. The result in \textbf{[1A]} is similar to \textbf{[2F]} even though there is no ASR adaptation and no verbatim transcripts.
It is interesting to observe that while the model can learn to process entities with just their corresponding values in spoken order, performance drops significantly when the entities and their values are reordered in \textbf{[2A]}. However, previous work has shown that attention based encoder-decoder models can perform surprisingly well in this case~\cite{kuo2020end}.

\subsection{SLU models in the \textbf{[TEXT]} setting}
In the \textbf{[TEXT]} setting, unlike the other two settings described above, we assume that transcripts and semantic labels are available but corresponding audio is absent. To circumvent the lack of real speech, as in~\cite{huang2020leveraging,lugosch2020using}, 
we use a TTS system \cite{Kons2019} to synthesize speech from the transcripts and then use the speech to perform SLU training.
Given that synthesized speech can be quite different from the test audio, we experiment with various methods to build SLU systems. In the first set of experiments, we adapt just the prediction and joint networks (PR+Joint) while keeping the transcription network fixed.
In Tables 3, 4 and 5, experiments \textbf{[1T]},  \textbf{[5T]},  \textbf{[9T]}, (marked with a $*$) are all oracle experiments training on real audio to compare to TTS data. 
We also evaluate the effect of using single speaker TTS data (Adapt-SS) versus multi-speaker TTS data (Adapt-MS). Experiments \textbf{[1T]} through \textbf{[4T]} in Table \ref{tab:entity_atis} are various entity recognition experiments in this setting. 
It is interesting that adapting just the prediction and joint networks (\textbf{[2T]}) on single speaker TTS data can be effective to create an SLU system, producing similar results as real speech data  (\textbf{[1T]}). More gains can, however, be obtained if the entire RNN-T model is adapted (ALL). In this case, multiple TTS speakers yielded better results than a single TTS speaker (compare \textbf{[3T]} to \textbf{[4T]}; Table \ref{tab:entity_atis}). 

\begin{table}[t]
    \caption{SLU intent recognition performance in various use case settings on the Call Center data set}
    \vspace{-2mm}
	\centering
	\begin{tabular}{@{}ccc@{}} \toprule
	\multicolumn{1}{c}{} &  \multicolumn{1}{c}{\bf Model Building} & \multicolumn{1}{c}{\bf Int. (Acc\%)} \\ 
	\midrule
	\textbf{[7A]} & Attention-based model\cite{kuo2020end} & 88.4 \\
	\textbf{[8A]} & ASR PT $\rightarrow$ SLU Adapt & 87.4 \\
	\textbf{[9A]} & ASR PT $\rightarrow$ \textit{ASR adapt} $\rightarrow$ SLU Adapt$^*$ & 87.7 \\
	\textbf{[10A]} & ASR+POS PT $\rightarrow$ SLU Adapt & 87.2 \\
	\midrule
	\textbf{[9T]} & ASR PT $\rightarrow$ SLU Adapt-Real (PR+Joint)$^*$  &   66.9 \\
	\textbf{[10T]} &ASR PT $\rightarrow$ SLU Adapt-SS (PR+Joint) & 60.3 \\
	\textbf{[11T]} & ASR PT  $\rightarrow$ SLU Adapt-SS (ALL) &    54.4 \\
	\textbf{[12T]} & ASR PT  $\rightarrow$ SLU Adapt-MS (ALL) &    83.5 \\
	\bottomrule
	\end{tabular}
	\label{tab:intent_cc}
\end{table}

We repeat similar experiments with synthetic speech on intent recognition  on the ATIS task in Table \ref{tab:intent_atis}. Similar trends are observed in this case as well (see results \textbf{[5T]} to \textbf{[8T]}; Table \ref{tab:intent_atis}). Given that the ATIS corpus was collected in  clean recording conditions, perhaps the TTS speech data matches well acoustically with the corpus data. 
Unlike the original data which contains significant portions of silences, the TTS data is more controlled, which may actually improve RNN-T modeling performance. Table~\ref{tab:intent_cc} shows corresponding results for the Call Center data set.  One significant difference is that adapting the entire network on a single  speaker TTS data is detrimental (\textbf{[11T]}). When multi-speaker TTS is used, adapting the entire network (\textbf{[12T]}) gives a good intent accuracy (83.5\%), but there is still a 4\% gap with using real speech data (\textbf{[8A]}) due to acoustic mismatch. Overall, these experiments show that useful SLU systems can be bootstrapped with synthesized speech data, with better performance when the TTS generated data is diverse and acoustically matches the test data.

\section{Conclusions}
\label{sec:conclusions}

In this paper we have developed state-of-the-art RNN-T models in various SLU settings. Our experiments show that although the amount of pre-training data is not critical, a pre-trained model is essential to construct these E2E models. 
While there is not a strong case for creating pre-trained models with pseudo-SLU labels like POS tags, there is evidence that SLU models can benefit from ASR adaptation, prior to SLU training.  Overall, the ASR accuracy of the pre-trained ASR model used to initialize domain-specific SLU training has relatively little effect on the final SLU performance, even when verbatim transcripts are unavailable during SLU training.  Remarkably, this is true even when each utterance is associated with only a single intent label without any word transcripts.  In a transcript only setting, TTS generated data is effective in place of real audio data to bootstrap an RNN-T based SLU model.

\vfill
\pagebreak
\bibliographystyle{IEEEbib}
\bibliography{strings,refs}

\end{document}